\documentclass{apple}
\usepackage{amsmath}
\usepackage{enumerate} 
\usepackage{algorithm}
\usepackage{amsfonts}
\usepackage{amsthm}
\usepackage{newtxtt}
\usepackage{diagbox} 
\usepackage{colortbl}
\usepackage{amssymb}
\usepackage{xspace}
\usepackage{wrapfig}
\usepackage{adjustbox}
\usepackage{tabularx}
\usepackage{booktabs}
\usepackage{mathtools}
\usepackage{tikz}
\usepackage{enumitem}
\usepackage{silence}
\usepackage{dsfont}
\usepackage[table]{xcolor}
\usepackage[dvipsnames]{xcolor}
\usepackage{multirow}
\usepackage{makecell}
\usepackage{xfakebold}
\usepackage{mystyle}
\usepackage{minted}
\usepackage{fancyvrb}
\usepackage{bbm}

\DeclareMathAlphabet{\mathmybb}{U}{bbold}{m}{n}

\definecolor{textgray}{HTML}{6E6E73}
\usetikzlibrary{positioning, calc}
\usetikzlibrary{decorations.pathmorphing}

\makeatletter
\patchcmd{\wrong@fontshape}{\@gobbletwo}{}{}{}
\makeatother
\makeatletter
\AtBeginDocument{%
  \urlstyle{sf}
  
}
\makeatother

\numberwithin{equation}{section} 
\tcbuselibrary{minted}
\setminted[python]{
  linenos,
  breaklines,
  fontsize=\footnotesize,
  xleftmargin=2em
}

\definecolor{light}{RGB}{125, 125, 125}
\crefname{tcb@cnt@pbox}{code}{code}
\Crefname{tcb@cnt@pbox}{Code}{Code}
\crefname{assumption}{assumption}{assumption}
\Crefname{assumption}{Assumption}{Assumptions}

\newtcolorbox[auto counter]{pbox}[2][]{
  colback=white,
  title=Code~\thetcbcounter: #2,
  #1,fonttitle=\sffamily,
  fontupper=\sffamily,
  arc=2pt,
  colframe=bgcolor,
  coltitle=fgcolor,
  colbacktitle=bgcolor,
  toptitle=0.25cm,
  bottomtitle=0.125cm
}

\title{Learning to Reason as Action Abstractions with Scalable Mid-Training RL}

\author[1,2,*]{Shenao Zhang}
\author[1]{Donghan Yu}
\author[1]{Yihao Feng}
\author[1,3,*]{Bowen Jin}
\author[2]{Zhaoran Wang}
\author[*,\dagger]{John Peebles}
\author[1,\dagger]{Zirui Wang}

\affiliation[1]{Apple}
\affiliation[2]{Northwestern University}
\affiliation[3]{UIUC}
\affiliation[*]{Work done at Apple}
\affiliation[\dagger]{Equal advising}

\abstract{
Large language models excel with reinforcement learning (RL), but fully unlocking this potential requires a mid-training stage. An effective mid-training phase should identify a compact set of useful actions and enable fast selection among them through online RL. We formalize this intuition by presenting the first theoretical result on how mid-training shapes post-training: it characterizes an action subspace that minimizes both the value approximation error from pruning and the RL error during subsequent planning. Our analysis reveals two key determinants of mid-training effectiveness: pruning efficiency, which shapes the prior of the initial RL policy, and its impact on RL convergence, which governs the extent to which that policy can be improved via online interactions. These results suggest that mid-training is most effective when the decision space is compact and the effective horizon is short, highlighting the importance of operating in the space of action abstractions rather than primitive actions. Building on these insights, we propose Reasoning as Action Abstractions (RA3), a scalable mid-training algorithm. Specifically, we derive a sequential variational lower bound and optimize it by iteratively discovering temporally-consistent latent structures via RL, followed by fine-tuning on the bootstrapped data. Experiments on code generation tasks demonstrate the effectiveness of our approach. Across multiple base models, RA3 improves the average performance on HumanEval and MBPP by 8 and 4 points over the base model and the next-token prediction baseline. Furthermore, RA3 achieves faster convergence and higher asymptotic performance in RLVR on HumanEval+, MBPP+, LiveCodeBench, and Codeforces.
}

\metadata[Correspondence]{\sffamily Shenao Zhang \url{shenao@u.northwestern.edu}; John Peebles \url{papers@johnpeebles.com}; Zirui Wang \url{ziruiw@apple.com}}
\date{\sffamily September 30, 2025}

\begin{document}

\maketitle

\section{Introduction}
The potential of reinforcement learning (RL) as a universal policy-improvement operator has been demonstrated with remarkable success in training large language models (LLMs), spanning applications in preference optimization \citep{ouyang2022training}, mathematics \citep{guo2025deepseek,zeng2025simplerl}, code generation \citep{yang2025qwen3,zeng2025acecoder}, and agentic tasks \citep{team2025kimi,zhou2025sweet}. A central factor behind these successes is the strengthened policy prior, often established through mid-training \citep{wang2025octothinker,su2025scaling}, which is continued pre-training on expert data sampled from the optimal policy. Despite its widespread use, the precise role of mid-training in shaping post-training RL remains poorly understood. This gap hinders the design of principled and effective mid-training algorithms. Existing heuristic metrics, such as the performance or entropy of the initial RL policy, offer only indirect signals and provide no guarantee of improved downstream outcomes.

An ideal mid-training algorithm should extract from finite expert demonstrations a set of useful actions sufficient for all tasks, and enable efficient selection among them during RL. We formalize this by presenting the first theoretical analysis of how mid-training shapes post-training RL. Specifically, mid-training should identify an action subspace that simultaneously minimizes (1) the approximation error induced by pruning the action space, and (2) the post-training RL error incurred when planning within the pruned space. Corresponding to these two objectives, we disclose two key factors that determine the effectiveness of mid-training: the efficiency of pruning the decision space and its impact on the convergence of subsequent RL. The first factor governs the learned prior encoded in the initial RL policy, while the second dictates the policy’s potential to be improved through online interactions. Our results show that pruning efficiency is inversely related to the cardinality of the smallest near-optimal action subset, and that post-training RL converges faster when actions are temporally extended. Together, these findings suggest that mid-training should operate in the space of action abstractions rather than primitive actions. Intuitively, learning high-level, transferable ``skills” benefits from a compact set of decisions and a reduced effective horizon, making pruning more efficient and planning more tractable.

To uncover this action hierarchy, we derive a temporal variational lower bound for the next-token prediction (NTP) objective. This bound can be optimized via an iterative expectation–maximization procedure, involving a self-supervised RL step that treats the log-likelihood of expert actions as reward to discover the hidden latent sequence, and a supervised fine-tuning step on the bootstrapped data. With an appropriate latent prior, the KL divergence enforces temporal consistency, ensuring that the latents function as coherent action abstractions. These design choices yield a scalable mid-training algorithm, Reasoning as Action Abstractions (RA3), where the KL penalty naturally regulates the need for rational rollouts, thereby controlling computational cost.

We evaluate RA3 on code generation tasks using Qwen and Llama base models ranging in size from $1$B to $8$B. The mid-training corpus consists of $3.5$M code snippets totaling $1$B tokens. Our experiments show that fine-tuning on data bootstrapped with action abstractions substantially reduces cross-entropy loss and improves performance across multiple benchmarks, including HumanEval, MBPP, and their extended variants. On average, RA3 achieves a 4-point improvement over NTP and an 8-point improvement over the base models. Furthermore, RA3 accelerates RLVR convergence and attains higher asymptotic performance on HumanEval+, MBPP+, LiveCodeBench, and CodeForces. These results underscore the scalability and effectiveness of learning action abstractions during mid-training.
\section{Background}\label{sec_bg}
\paragraph{Imitation Learning.} A task $\mathcal{M}=(\mathcal{S}, \mathcal{A}, R, \gamma)$ is an MDP with state space $\mathcal{S}$, (primitive) action space $\mathcal{A}$, reward function $R$, and discount factor $\gamma<1$. In the language setting, a state corresponds to the context of all previously generated tokens, and each primitive action is a single token. State transitions are either deterministic, by appending the new action tokens to the context, or governed by the external environment. An expert policy is the policy that maximizes the expected return: 
\$
\pi_E \in\argmax_\pi\EE_{a_t\sim\pi}\Bigl[R(s_t, a_t) + \gamma V_\mathcal{M}^\pi(s_{t+1})\Bigr] = \EE_{a_t\sim\pi}\Bigl[\sum\nolimits_{t} \gamma^t R(s_t, a_t)\Bigr].
\$ 
It is worth noting that when the task is inherently solvable within a single step, such as math problems where $r(s_0, a_\text{gt})=1$, the expert policy deterministically outputs the ground-truth answer $a_\text{gt}$ at $s_0$, i.e., $\pi_E(a_\text{gt}|s_0) = 1$, to maximize the return. However, since most mid-training math data includes explicit human reasoning before reaching $a_\text{gt}$, we instead focus on the multi-step decision-making domains, such as code generation and agentic tasks, where expert trajectories are sequences of actions. 

Next-token prediction (NTP) during mid-training can be viewed as imitation learning on an offline expert dataset $\mathcal{D}_E$, collected by rolling out $\pi_E$ on sampled tasks $\mathcal{M}\sim p(\mathcal{M})$. Its objective is to maximize the conditional log-likelihood:
\#\label{eq_ntp}
\mathcal{J}_{\text{NTP}}(\pi) = \EE_{(s_{0:T}, a_{0:T})\sim\mathcal{D}_E}\Bigl[\log \pi(a_{0:T}\mid s_{0:T})\Bigr] = \EE_{\mathcal{D}_E}\Biggl[\sum_{t=0}^T \log \pi(a_t\mid s_t)\Biggr],
\#
where $s_t\in\mathcal{S}$, $a_t\in\mathcal{A}$, $T$ is the sequence length of an expert demonstration, $\pi$ is the training policy, and $s_0$ is the beginning of the sentence (BOS) token. The formula in \eqref{eq_ntp} applies not only to token-level actions but also to coarser-grained actions, such as sentence-level actions.

NTP is widely used across different stages of LLM training—pre-training, continued pre-training (or mid-training when the goal is to acquire reasoning foundations before RLVR), and supervised fine-tuning. In this work, we focus primarily on mid-training, the second stage in the three-stage pipeline: pre-training, mid-training, and RLVR post-training.

\paragraph{RL with Verifiable Reward.} The objective of post-training RL is to maximize the expected return. In RLVR, a common setup is to formulate the problem as a single-step MDP with a binary, outcome-based reward, defined as $r(s, o)=\text{verifier}(s, o)$ to check whether the model response $o$ matches the ground-truth answer associated with the prompt question $s$. We adopt Group Relative Policy Optimization (GRPO) \citep{shao2024deepseekmath,guo2025deepseek} as our default RLVR algorithm in experiments. Its objective is
\$
\mathcal{J}_{\text{GRPO}}(\pi) = \frac{1}{G}\sum_{i=1}^G\biggl(\min\biggl(\frac{\pi(o_i\mid s)}{\pi_{\text{old}}(o_i\mid s)}A_i, \text{clip}\biggl(\frac{\pi(o_i\mid s)}{\pi_{\text{old}}(o_i\mid s)}, 1\pm\epsilon\biggr)A_i\biggr) - \beta\mathcal{D}_{\text{KL}}(\pi, \pi_{\text{ref}})\biggr),
\$
where $o_i\sim\pi_{\text{old}}(\cdot| s)$, $\epsilon, \beta$ are hyperparameters, and the advantage is calculated within the group $G$:
\#\label{eq_grpo_adv}
A_i = \bigl(r(s, o_i) - \text{mean}(\{r(s, o_i)\}_{i=1}^G)\bigr) / \bigl(\text{std}(\{r(s, o_i)\}_{i=1}^G)\bigr).
\#
\section{How Mid-Training Shapes Post-Training RL}\label{sec_analysis}
The goal of post-training RL is to minimize regret:
\#\label{eq_postrl}
\min_\pi\EE_{\mathcal{M}\sim p(\mathcal{M})}[V^*_{\mathcal{M}}(s_0) - V^{\pi}_{\mathcal{M}}(s_0)],
\#
that is, to learn $\pi$ that is near-optimal across tasks $\mathcal{M}\sim p(\mathcal{M})$. In what follows, to distinguish from the primitive action space $\mathcal{A}$ introduced in Section \ref{sec_bg}, we consider a unified action space $\mathcal{Z}$, which generalizes $\mathcal{A}$ by including action abstractions that will be defined later. 

The major role of mid-training is to identify a subspace of useful actions to avoid exploring the vast language space during RL. To formalize this, we first define the quality of an action subset. Consider $\mathcal{M}_{\mathcal{Z}'}=(\mathcal{S}, \mathcal{Z}', R, \gamma)$, which is identical to $\mathcal{M}$ except that the action space is restricted to a subset $\mathcal{Z}'\subset\mathcal{Z}$. We say $\mathcal{Z}'$ is $\epsilon$-optimal for $\mathcal{M}$ if $\epsilon$-optimal policies can be constructed using only the actions in $\mathcal{Z}'$. Formally,
\begin{definition}[$\epsilon$-Optimal Task Action Subset]\label{def_no_action}
An action subset $\mathcal{Z}'\subseteq\mathcal{Z}$ is called $\epsilon$-optimal for task $\mathcal{M}$ if the optimal values in $\mathcal{M}$ and $\mathcal{M}_{\mathcal{Z}'}$ satisfy $\Delta(\mathcal{M}, \mathcal{Z}') := V_\mathcal{M}^{*}(s_0) - V_{\mathcal{M}_{\mathcal{Z}'}}^{*}(s_0)\leq \epsilon$.
\end{definition}

This definition allows us to connect mid-training with post-training RL through the following decomposition:
\begin{lemma}[Regret Decomposition]\label{lemma_r}
For any $\mathcal{Z}'\subseteq\mathcal{Z}$, the post-training RL regret in \eqref{eq_postrl} satisfies
\$
\EE_{\mathcal{M}\sim p(\mathcal{M})}\bigl[V^*_{\mathcal{M}}(s_0) - V^{\pi}_{\mathcal{M}}(s_0)\bigr] = \underbrace{\EE_{\mathcal{M}\sim p(\mathcal{M})}[\Delta(\mathcal{M}, \mathcal{Z}')]}_{\text{action-set pruning error}} + \underbrace{\EE_{\mathcal{M}\sim p(\mathcal{M})}[V^*_{\mathcal{M}_{\mathcal{Z}'}}(s_0) - V^{\pi}_{\mathcal{M}_{\mathcal{Z}'}}(s_0)]}_{\text{post-training RL error}}.
\$
\end{lemma}
Lemma \ref{lemma_r} indicates that mid-training should identify an action subspace $\hat{\mathcal{Z}}$ that minimizes two sources of error: (1) the approximation error from pruning the full action space $\mathcal{Z}$ down to $\hat{\mathcal{Z}}$, and (2) the post-training RL error incurred when planning within $\hat{\mathcal{Z}}$.

Corresponding to these two terms, we will reveal two key factors that govern the effectiveness of mid-training: pruning efficiency, which determines the initial RL policy prior, and the impact on RL convergence, which determines how efficiently the policy can be further improved through online interactions.

\subsection{Mid-Training Acquires Policy Priors via Action Space Pruning}
A central metric of mid-training pruning is how efficiently the algorithm can eliminate useless actions, since it directly determines the quality of the resulting action space given a fixed number of expert demonstrations. To formalize this, we introduce the following notion.

\begin{definition}[Minimal Size of $\epsilon$-Optimal Action Subset]
$\overline{\mathcal{Z}}_\epsilon\subseteq\mathcal{Z}$ is an $\epsilon$-optimal action subset if it is $\epsilon$-optimal for all tasks. Let $|\overline{\mathcal{Z}}_\epsilon|$ denote the minimal size of such a subset.
\end{definition}

Notably, $|\overline{\mathcal{Z}}_\epsilon|$ is always finite for any $\epsilon\geq0$ because even when $\epsilon=0$, we may still take $\overline{\mathcal{Z}}_{0}=\mathcal{Z}$ as $\Delta(\mathcal{M},\mathcal{Z})=0$ for all $\mathcal{M}$. In order words, in the worst case, no pruning occurs and all the actions in the finite $\mathcal{Z}$ are retained. We now present our result on pruning efficiency during mid-training.

\begin{theorem}[Pruning Efficiency]\label{thm_expert_size}
Assume the rewards are within the range of $[0, 1]$. Denote $|\mathcal{D}_E|$ as the number of expert demonstrations during mid-training and $\hat{\mathcal{Z}}$ as the resulting pruned action space. If
\$
|\mathcal{D}_E|=\Theta\Bigl(|\overline{\mathcal{Z}}_\epsilon|\log(|\mathcal{Z}| / \delta) / \sigma\Bigr),
\$
then with probability at least $1-\delta$, the pruning error in Lemma \ref{lemma_r} satisfies $\EE_{\mathcal{M}\sim p(\mathcal{M})}[\Delta(\mathcal{M}, \hat{\mathcal{Z}})] \leq \epsilon(1-\sigma)+\sigma/(1-\gamma)$.
\end{theorem}

Theorem \ref{thm_expert_size} shows that the number of expert samples required to prune suboptimal actions decreases as both $|\overline{\mathcal{Z}}_\epsilon|$ and $|\mathcal{Z}|$ shrink. With a higher pruning efficiency, i.e., a smaller $|\overline{\mathcal{Z}}_\epsilon|$ and $|\mathcal{Z}|$, the probability that suboptimal actions survive mid-training decreases, leading to a smaller pruning error. This result highlights the importance of a compact decision space and motivates defining $\mathcal{Z}$ as a space of action abstractions instead of primitive actions $\mathcal{A}$.

Action abstractions are defined analogously to Markov options \citep{puterman1994markov,sutton1999between,precup2000temporal}, representing the abstraction of temporally-extended primitive actions. Each $z \in \mathcal{Z}$ corresponds to a high-level intention that executes a sequence $a_i, \dots, a_{i+\tau}$, with duration $\tau \sim p(\cdot \mid s, z)$. Notably, the above result for $\mathcal{Z}$ also applies to $\mathcal{A}$ as a special instantiation of $\mathcal{Z}$, with $\tau=1$ and $a_j=z_j$. 

Compared to the space of primitive actions $\mathcal{A}$, the space of action abstraction has a substantially smaller size of $|\mathcal{Z}|$ and $|\overline{\mathcal{Z}}_\epsilon|$, since each action abstraction corresponds to a transferable skill spanning multiple tasks. Theorem \ref{thm_expert_size} implies that learning directly at the primitive action level demands significantly more expert data, while temporal abstractions enable mid-training to more efficiently approximate a near-optimal action set and provide a stronger policy prior, thereby reducing the burden on post-training RL.

\subsection{Mid-Training Accelerates RL Convergence}
The analysis above focuses on pruning error while holding the post-training RL error fixed. However, even with zero pruning error (e.g., by retaining the full primitive action space $\hat{\mathcal{Z}}=\mathcal{A}$), RL may still be highly inefficient for long-horizon tasks due to the vast action space. We now analyze how mid-training influences the convergence of online RL. Our result is based on policy iteration due to its simplicity, which iteratively updates the policy to optimality w.r.t. the policy value.

\begin{theorem}[RL Convergence Rate]\label{thm_conv}
After at most $N=\mathcal{O}((|\mathcal{S}||\hat{\mathcal{Z}}| - |\mathcal{S}|) / (1 - \overline{\gamma}))$ iterations, the policy achieves optimality in the pruned action space $\hat{\mathcal{Z}}$, where $\overline{\gamma} = \sup_{s, z}\EE[\gamma^\tau | s, z]\leq \gamma$.
\end{theorem}

The result reveals that reasoning structures acquired during mid-training affect convergence through both the pruned action space size and the duration $\tau$ of the temporally-extended actions. Longer actions reduce $\overline{\gamma}$, leading to faster RL convergence than NTP-style mid-training, where $\tau=1$ and $\overline{\gamma}=\gamma$. This makes intuitive sense as each Bellman backup jumps across $\tau$ steps in one shot, which shortens the effective planning horizon and shrinks the error faster per iteration. Similar dependency on the effective planning horizon also appears in bounds for broader RL algorithms, such as policy gradient \citep{agarwal2021theory,zhang2023model}, though we omit them due to additional assumptions. Moreover, the number of suboptimal actions that RL needs to eliminate during online interaction reduces with a more compact action space $\hat{\mathcal{Z}}$.

\begin{figure}[h]
    \centering
    \begin{minipage}[b]{0.185\linewidth}
        \centering
        \includegraphics[width=1.0\linewidth]{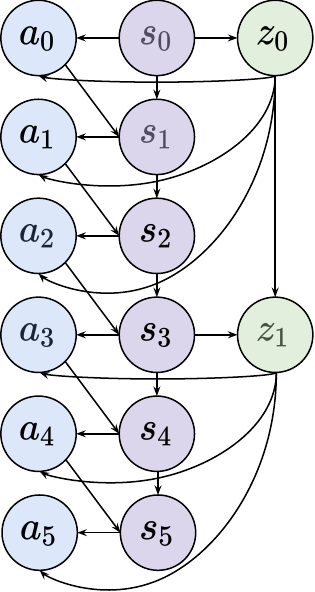}
    \end{minipage}
    \begin{minipage}[b]{0.23\linewidth}
        \centering
        \includegraphics[width=0.85\linewidth]{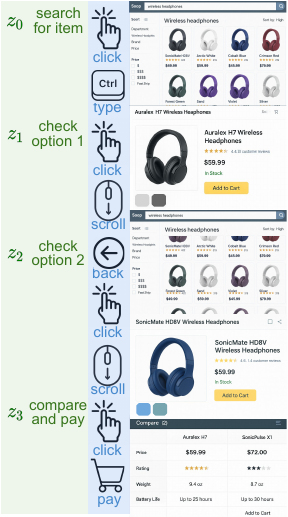}
    \end{minipage}
    \begin{minipage}[b]{0.53\linewidth}
        \centering
        \includegraphics[width=1\linewidth]{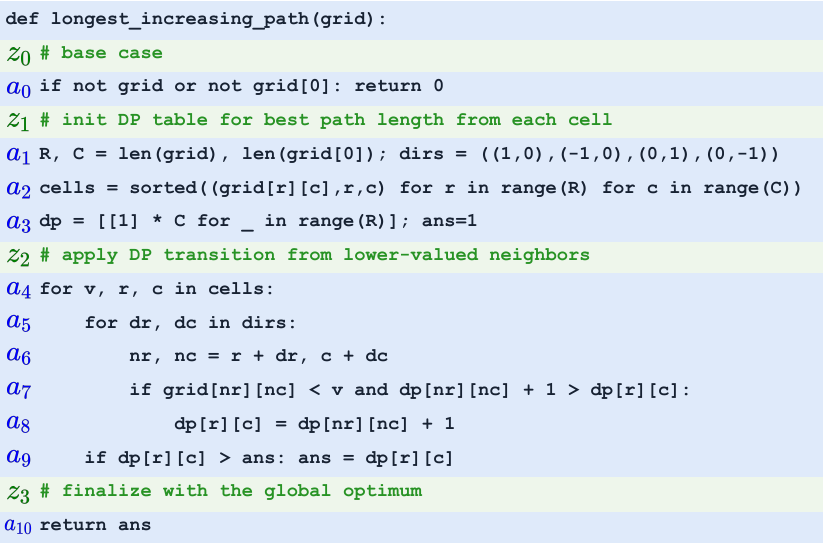}
    \end{minipage}
\vspace{-0.1cm}
\caption{\textbf{(Left)}: The probabilistic graphical model of the action hierarchy. \textbf{(Middle \& Right)}: Examples of primitive actions in expert demonstrations (blue) and the hidden high-level temporal abstractions (green), in web agent and code generation domains, respectively.}
\label{fig_pgm}
\end{figure}

Intuitively, there are high-level ``skills" that are shared across tasks. Leveraging these skills yields a compact decision set and shortens the planning horizon, which makes pruning more efficient and planning more tractable. We illustrate this intuition with two examples in Figure \ref{fig_pgm}.

\section{From Primitive Actions to Temporal Abstractions}
In Section \ref{sec_analysis}, we analyze the benefits of learning temporal action abstractions during mid-training from two perspectives: their efficiency in pruning the action space and their ability to accelerate subsequent RL. In what follows, we present a principled method for extracting temporal abstractions from primitive actions in mid-training expert demonstrations.

\subsection{Temporal Variational Bound}
We begin by seeking an alternative way to maximize the likelihood beyond predicting the next tokens. Specifically, we derive a sequential Evidence Lower Bound (ELBO) for the NTP objective:
\begin{theorem}[Temporal ELBO]\label{thm_elbo}
The next-token prediction objective in \eqref{eq_ntp} is lower bounded by
\small
\$
\mathcal{J}_{\text{NTP}}(\pi) \geq \mathcal{J}(\pi, q) = \EE_{(s_{0:T}, a_{0:T})\sim\mathcal{D}_E, z_t\sim q}\Biggl[\sum_{t=0}^{T}\log \pi(a_t| s_t, z_t) - \mathcal{D}_{\text{KL}}\Bigl(q(z_t| s_t, z_{0:t-1}) \,\,||\,\, p(z_t| s_t, z_{0:t-1})\Bigr)\Biggr],
\$
where $p(z_t| s_t, z_{t-1})$ is the prior distribution of $z_t$.
\end{theorem}

The ELBO introduces a sequence of latents $z_{0:T}$ to model the observed primitive actions $a_{0:T}$. Intuitively, these latents represent hidden thoughts or intentions behind the expert decisions $a_{0:T}$ that are not explicitly available in the mid-training data. Amortized variational inference employs a parametric variational posterior $q(z_t|s_t, z_{0:t-1})$ as a tractable surrogate for the intractable true posterior over $z$. Maximizing the ELBO amounts to estimating the latent trajectories $z_{0:T}$ that rationalize expert behavior, thereby tightening the bound on the marginal likelihood $p(a_{0:T} \mid s_{0:T})$.

We maximize the ELBO by alternatively optimizing $q$ and $\pi$, in an Expectation-Maximization (EM) manner. In each EM iteration $i$, the E step fixes $\pi_i$ and updates $q$:
\#\label{eq_e}
q_i = \argmax_q \mathcal{J}(\pi_i, q) = \argmax_q \EE_{\mathcal{D}_E, z_t\sim q}\Biggl[\sum_{t=0}^{T}\underbrace{\log \pi_i(a_t\mid s_t, z_t) - \mathcal{D}_{\text{KL}}\bigl(q\,||\, p\bigr)}_{\text{RL reward at step } t}\Biggr].
\#
This corresponds to a $T$-horizon RL procedure where the per-step reward is the log-likelihood of the observed expert actions, with a KL penalty that will be discussed later. Intuitively, \eqref{eq_e} encourages the latent sequence $z_{0:T}$ sampled from $q_i$ to ``explain" the expert decisions $a_{0:T}$.

The $M$-step then fixes the updated $q_i$ and optimizes $\pi$:
\#\label{eq_m}
\pi_{i+1} = \argmax_\pi \mathcal{J}(\pi, q_i) = \argmax_\pi\EE_{\mathcal{D}_E, z_t\sim q_i}\bigl[\log \pi(a_t\mid s_t, z_t) \bigr],
\#
which is simply imitation learning on expert trajectories bootstrapped with the inferred latents $z_t$.

\subsection{Temporally-Consistent Latents as Action Abstractions}
We have reformulated likelihood maximization on primitive actions $a_{0:T}$ as optimizing an ELBO that learns from the bootstrapped sequence with latent trajectories sampled from the variational posterior $q(z_t|s_t, z_{0:t-1})$. 
Recall that our analysis in Section \ref{sec_analysis} reveals the benefits of compact sets of high-level action abstractions. We will show in the following how this can be achieved by properly designing the latent prior.

The latent $z_t$ in the ELBO is defined per-step for every $t\in[0, T]$. To represent temporally extended abstractions spanning $\tau \sim p(\cdot \mid s_t, z_t)$ steps, we require $z_t = z_{t+1} = \cdots = z_{t+\tau}$\footnote{The notation here deviates slightly from Section \ref{sec_bg}, where all $\tau$ identical latents are written as a single $z$.}. This is enforced by defining the prior $p(z_{t+1}| s_{t+1}, z_{t})$ in Theorem \ref{thm_elbo} to have a large probability mass on $z_{t}$, and uniformly distributed at all other positions:
\#\label{eq_prior}
p(z_t\mid s_t, z_{t-1}) = \alpha\delta(z_{t-1}) + (1 - \alpha)U(z_t),
\#
where $\alpha\in[0, 1]$ is a hyperparameter, $\delta(\cdot)$ is the Dirac delta function, and $U(\cdot)$ is the uniform distribution over the finite set $\mathcal{Z}$. The delta mass helps preserve a temporally-consistent latent as a high-level action abstraction (reusing $z_{t-1}$), while the uniform term promotes diversity across latents.

\section{RA3: A Scalable Mid-Training Algorithm}
In the context of LLMs, the action abstraction $z$ serves a similar role as a rational or intention, and the primitive action $a$ corresponds to the actual token or operation in the environment. Taking code generation as an example, $z_t$ might capture reasoning steps that precede writing the next code block $a_{t+\tau}$.

In addition to the theoretical benefits discussed in Section \ref{sec_analysis}, the temporal consistency of the latents also enables scaling to mid-training-sized data. Directly generating rationals $z_{0:T}$ requires $T$ rollouts, proportional to the total number of tokens in the data. This prohibits us from making full use of the high-quality data in the mid-training corpus, which typically contains billions of tokens. Fortunately, the temporal consistency of latents rescues us from resampling rollout $z_t$ at each timestep $t$, as it remains fixed for $\tau$ steps.

To exploit this, we define two types of latents: $z=\text{\textless{}act\textgreater{}}$ and $z$ that begins with $\text{\textless{}think\textgreater{}}$. When the latent is temporally-consistent at timestep $t$, $a_t$ is generated directly without sampling new rationals since $z_t=z_{t-1}$ and $a_t$ shares the same high-level intention as $a_{t-1}$. For this reason, we use $z_t=\text{\textless{}act\textgreater{}}$ to indicate it is temporally-consistent. Rollouts terminate immediately upon sampling $\text{\textless{}act\textgreater{}}$. BIn contrast, full rollouts are sampled only when $\langle \text{think} \rangle$ initiates a new rationale, significantly reducing RL inference cost. This design modifies the KL term in the ELBO as follows.

\begin{proposition}\label{prop_1}
With the prior defined in \eqref{eq_prior}, the KL term in Theorem \ref{thm_elbo} satisfies
\small
\$
\mathcal{D}_{\text{KL}}\Bigl(q(z_t| s_t, z_{0:t-1}) \,\,||\,\, p(z_t| s_t, z_{0:t-1})\Bigr) = \mathcal{D}_{\text{KL}}\Bigl(\text{Bern}(q_{\text{act}}) \,\,||\,\, \text{Bern}(\alpha)\Bigl) - \mathcal{H}\Bigl(q(z_t| s_t, z_{0:t-1})\Bigr) + \text{const},
\$
where $q_{\text{act}} = q(z_t=\text{\textless{}act\textgreater{}}\mid s_t, z_{0:t-1})$, $\text{Bern}(\cdot)$ is the Bernoulli distribution, and
\small
\$
\mathcal{D}_{\text{KL}}\Bigl(\text{Bern}(q_{\text{act}}) \,\,||\,\, \text{Bern}(\alpha)\Bigl) = \EE_{z\sim q}\Bigl[\mathmybb{1}(z_t=\text{\textless{}act\textgreater{}})\log\frac{q_{\text{act}}}{\alpha} + \mathmybb{1}(z_t\neq\text{\textless{}act\textgreater{}})\log\frac{1 - q_{\text{act}}}{1 - \alpha}\Bigr].
\$
\end{proposition}
The KL decomposes into a Bernoulli KL regularizer and an entropy term. By setting $\alpha>q_{\text{act}}$, the KL discourages unnecessary thinking, since it assigns a larger penalty to $z_t\neq\text{\textless{}act\textgreater{}}$ than $z_t=\text{\textless{}act\textgreater{}}$. The penalty difference defines a threshold, guiding the model to generate new rationals only when they improve the log-likelihood reward by more than this threshold. In implementation, instead of tuning $\alpha$, we apply reward shaping and assign a fixed penalty to $z_t\neq \text{\textless{}act\textgreater{}}$. Proposition \ref{prop_1} also indicates that the additional training cost relative to NTP can be adjusted by $\alpha$. In the extreme case where $\alpha=1$, RA3 degenerates to NTP, since $\mathcal{D}_{\text{KL}}(\text{Bern}(q_{\text{act}}) \,||\,\text{Bern}(1))$ is infinite for all $q_{\text{act}}<1$, i.e., the $q$ policy receives an infinite penalty for generating any rationals. We provide the pseudocode in Algorithm \ref{alg_1}.

\vspace{-0.2cm}
\begin{algorithm}[H]
\caption{Reasoning as Action Abstractions (RA3) for Mid-Training}\label{alg_1}
\begin{algorithmic}[1]
\State \textbf{Input: } Base LLM $\pi_0$, mid-training dataset $\mathcal{D}_E$, penalty hyperparameter $c$.
\For{EM iteration $i$ in $1, 2, \cdots$}
\State Optimize $q_i=\argmax_q \EE_{\mathcal{D}_E^{e_i}, z_t\sim q}[\sum_{t=0}^{T}\log \pi_i(a_t| s_t, z_t) - c\mathmybb{1}(z_t\neq\text{\textless{}act\textgreater{}})]$ via RL.
\State Fine-tune $\pi_{i+1} = \argmax_\pi\EE_{\mathcal{D}_E^{m_i}, z_t\sim q_i}\bigl[\log \pi(a_t\mid s_t, z_t) \bigr]$ via NTP.
\EndFor
\end{algorithmic}
\end{algorithm}
\vspace{-0.2cm}

We enforce the two types of latents discussed above by incorporating a simple format reward in the RL step that assigns zero reward to the wrong format, which we omit in the pseudocode for clarity.
We optimize the $T$-horizon RL objective using policy gradient. Advantages are calculated in a way similar to \eqref{eq_grpo_adv} in GRPO: after sampling $G$ length-$T$ rollouts, we set the baselines as $b(s_{t'}) = \sum_{g=1}^G\sum_{t=t'}^T r_t^g/G$ that are independent of actions, and combine it with the state-action value to calculate the advantage at each step.
\section{Related Work}
\paragraph{LLM Mid-Training.} RL has long been utilized for training language models \citep{nguyen2017reinforcement,paulus2017deep,jaques2020human,ramamurthy2022reinforcement,ouyang2022training}. However, its potential as a universal policy-improvement operator has only recently been realized, as reasoning models learn to cast intermediate thoughts as actions and optimize them via RL \citep{guo2025deepseek,zeng2025simplerl,liu2025scaling,zhang2025beyond,team2025kimi}. This paradigm also informs the design of the policy prior. To obtain strong initial policies in terms of both performance and exploration diversity, mid-training \citep{xu2025phi,wang2025octothinker,su2025scaling} has become an essential step, which performs reasoning or agentic continued pre-training on high-quality expert data. Previous work that leverages next-token prediction during mid-training is an imitation learning process, where the data comes from rollouts of optimal expert policies, such as human demonstrations in device control and code writing \citep{rawles2023androidinthewild,huang2024opencoder,bai2024digirl}. Anchoring our findings, prior work observes that learning action hierarchies with abstraction-based reasoning outperforms training only on primitive actions alone \citep{xu2024aguvis,chen2024comments,wang2025opencua,xue2025simpletir}. For similar reasons, some mid-training datasets augment the expert data with synthetic reasoning distilled from frontier LLMs \citep{wang2025opencua,numina_math_datasets}. However, such distillation introduces distributional shift, and it remains unclear how much student models benefit compared to our method, which learns its own reasoning via RL. Moreover, RA3 avoids the high cost of reasoning distillation at scale. In fact, in the code generation domain of interest here, most mid-training datasets consist primarily of human-written code scraped from the internet, without costly reasoning relabeling.

\paragraph{Self-Supervised RL.} Optimizing the temporal variational bound involves a self-supervised RL step with log-prob of the expert action as reward \citep{zhong2025brite,ruan2025reasoning,zhou2025reinforcing,dong2025reinforcement,wang2025reverse}. Compared to \cite{zhong2025brite,ruan2025reasoning} that also uses EM-style updates, we are motivated to learn action abstractions with the ELBO derived for temporal sequences. Moreover, \cite{dong2025reinforcement} hand-crafts an entropy-based rule to determine reasoning positions and trains on only $4$k samples with instructed reasoning models. In contrast, RA3 is theoretically grounded, scales to mid-training setups, and enables the model to autonomously decide when to skip unnecessary reasoning through the temporal consistency of latents.

\paragraph{Markov Options.} RL based on options \citep{sutton1999between,precup2000temporal,bacon2017option} equips agents with varying courses of actions at extended time scales and learn in the MDP with them. It provides a natural form of hierarchical structure that facilitates learning in long-horizon tasks \citep{jong2008utility}. Our analysis of the decision space size is partly inspired by studies on option transfer in lifelong and multi-task RL \citep{brunskill2013sample,brunskill2014pac}. But unlike those works, which focus on option transferability, we are primarily concerned with designing mid-training algorithms for LLMs and understanding their impact on post-training RL.

\section{Experiments}
\paragraph{Experiment Setups.} We focus on Python code generation in our experiments. Primitive actions $a$ are defined at the granularity of a single line of code. For the two types of latents $z$, to avoid additional fine-tuning on special tokens, we remove the newline character \textbackslash n at the end of $a$ and set $\text{\textless{}act\textgreater{}}$=\textbackslash n, $\text{\textless{}think\textgreater{}}$=\textbackslash n\verb|#|. Thus, after line $a_t$, the model either outputs \textit{only} \textbackslash n before $a_{t+1}$ or generates a comment line as a high-level abstraction to guide code writing. The format reward is non-zero for the think action if it begins and ends with \textbackslash n and the first non-space token is $\verb|#|$. This design ensures that the reasoning bootstrapped data has the correct syntax and no additional filtering of the thinking sequences is needed. For the RL step of RA3, We replace the $T$-horizon objective with a $5$-step truncated return, attributing credit for action only to rewards within the next five steps. The RL optimization is implemented analogously to multi-turn RL: the code lines $a_t$ and reasoning comments $z_t$ alternate as turns. We adopt asynchronous rollout with the SGLang engine \citep{zheng2024sglang}, since batched inference often incurs idle time between turns. That is, batched rollouts must wait for all $z_t$ to be generated before proceeding to the next turn, which is particularly inefficient when temporal consistency is intended to minimize reasoning frequency.

We apply RA3 on three pre-trained models: Qwen-2.5-1.5B \citep{hui2024qwen2}, Llama-3.2-1B, and Llama-3.1-8B \citep{dubey2024llama}. The mid-training data is drawn from the Python subset of the continued pre-training corpus of \cite{huang2024opencoder}, containing $2.36$M high-quality internet code snippets ($834$M tokens) and $129$K code-only synthetic snippets ($120$M tokens). This leads to a total mid-training size of $3.5$M code snippets with $1$B tokens. Each EM iteration involves $400$ gradient updates, with the first $40$ being the RL policy gradient. A short warmup stage of $10$ RL gradient updates is applied without KL penalty, allowing the policy to initially focus on extracting the hidden rationales. The hyperparameters of the M step of RA3 is the same as NTP: batch size of $512$ with a learning rate of $2e-5$. The RL step of RA3 sets the maximum length of $z$ as $16$, with a sampling temperature of $1.0$ and a group size of $G=3$. The entropy coefficient is set to $0.001$ and we do not regularize the reference KL. The RL batch size is set to $1024$ with a learning rate of $2e-6$ and the penalty is set to $c=0.05$. AdamW optimizer is used \citep{loshchilov2017decoupled} for all algorithms.

\begin{wrapfigure}{r}{0.29\textwidth}
     \vspace{-0.4cm}
    \begin{minipage}[t]{0.29\textwidth}
        \centering
        \begin{minipage}[t]{\linewidth}
            \centering
            \includegraphics[width=\linewidth]{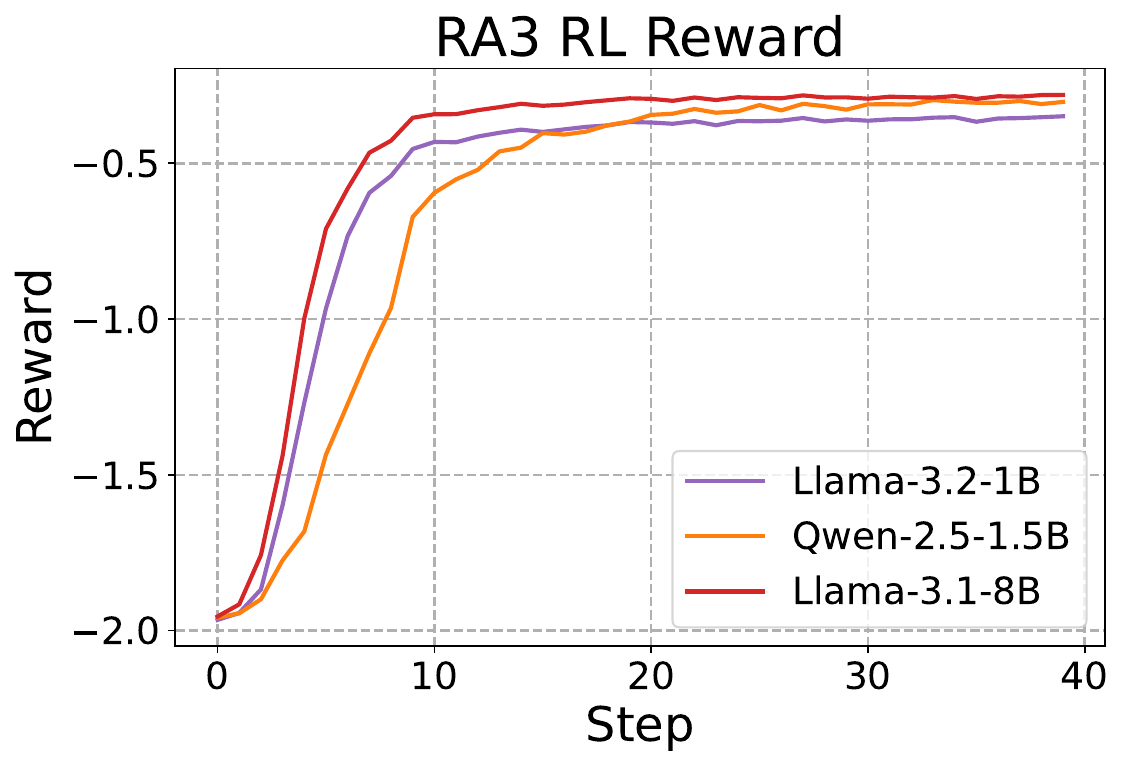}
        \end{minipage}
        \vspace{-0.72cm}
        \setlength{\belowcaptionskip}{-15pt}
        \caption{RL training reward curve.}
        \label{fig_rlcurve}
    \end{minipage}
\end{wrapfigure} 
\vspace{-0.1cm}
\paragraph{Mid-Training Results.} We begin by analyzing the results during the EM steps of RA3. The RL training reward in the first E step is shown in Figure \ref{fig_rlcurve}. We find that the model quickly learns to maximize the reward, meaning that most data and compute can be allocated to reasoning bootstrapping and supervised fine-tuning in the M step.

\begin{figure}[h]
    \centering
    \includegraphics[width=1.0\linewidth]{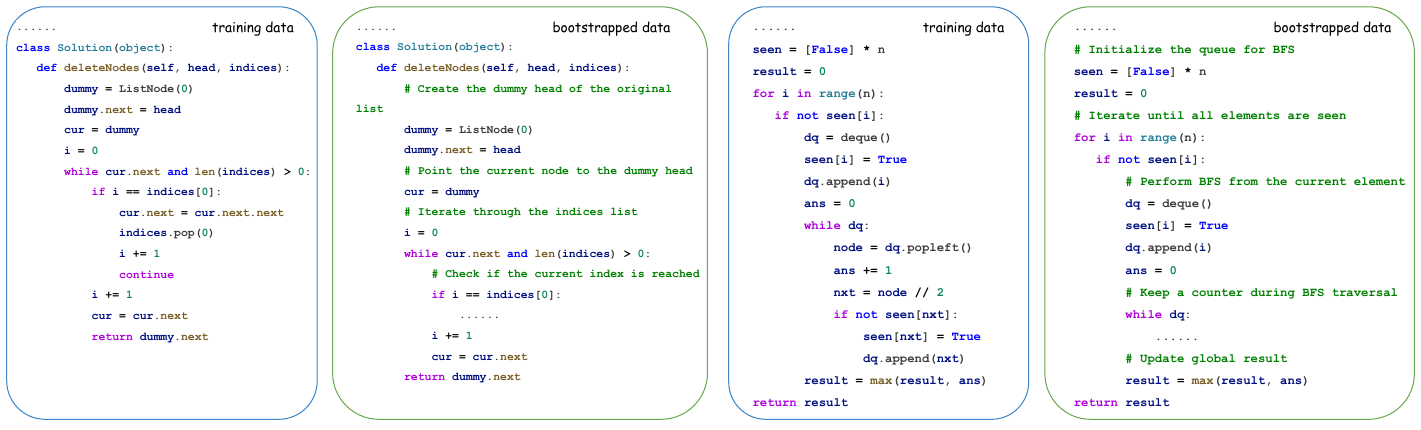}
    \vspace{-0.6cm}
\caption{Examples of the data from mid-training and after reasoning bootstrapping, where transferable skills, such as dummy head creation and BFS, are abstracted and incorporated into the data.}
\vspace{-0.2cm}
\label{fig_example}
\end{figure}

We also provide two pairs of examples of the training data before and after the RL steps in Figure \ref{fig_example}. After RL, the policy demonstrates the ability to extract high-level abstractions of expert behaviors embedded in the training data. These abstractions often correspond to transferable skills, such as dummy head creation and BFS, which in turn enhance the model’s generalizability.

\begin{figure}[htbp]
    \centering
    \begin{minipage}[b]{0.325\linewidth}
        \centering
        \includegraphics[width=1.0\linewidth]{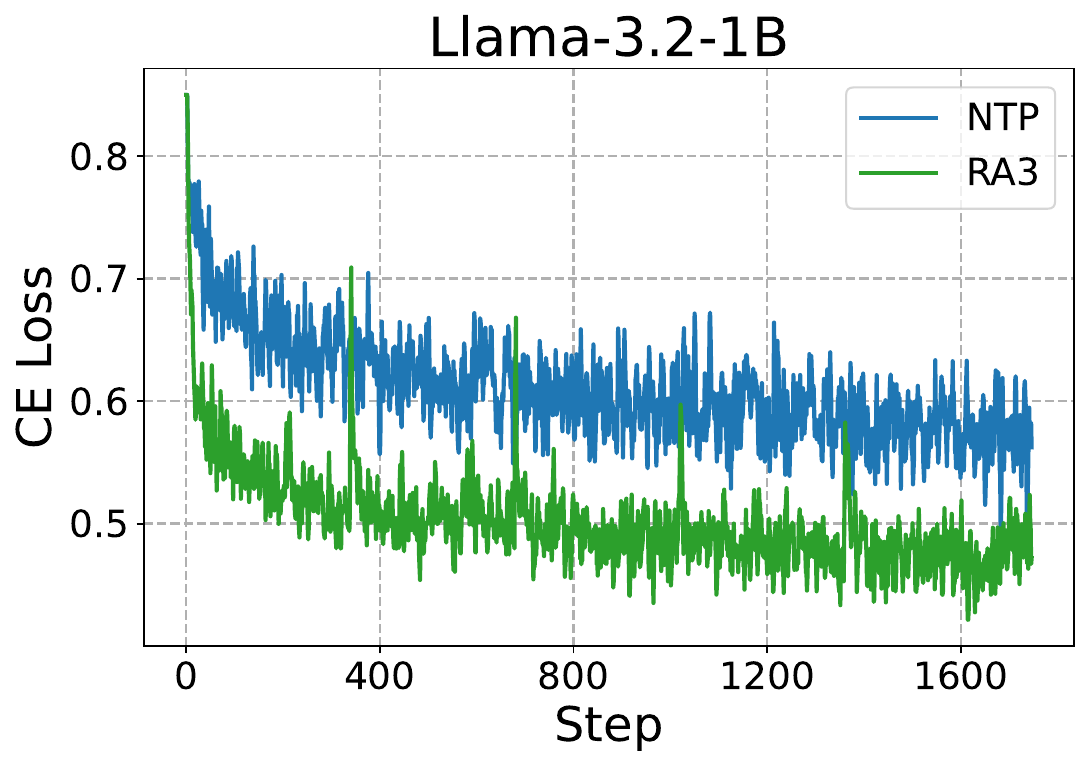}
    \end{minipage}
    \begin{minipage}[b]{0.325\linewidth}
        \centering
        \includegraphics[width=1.0\linewidth]{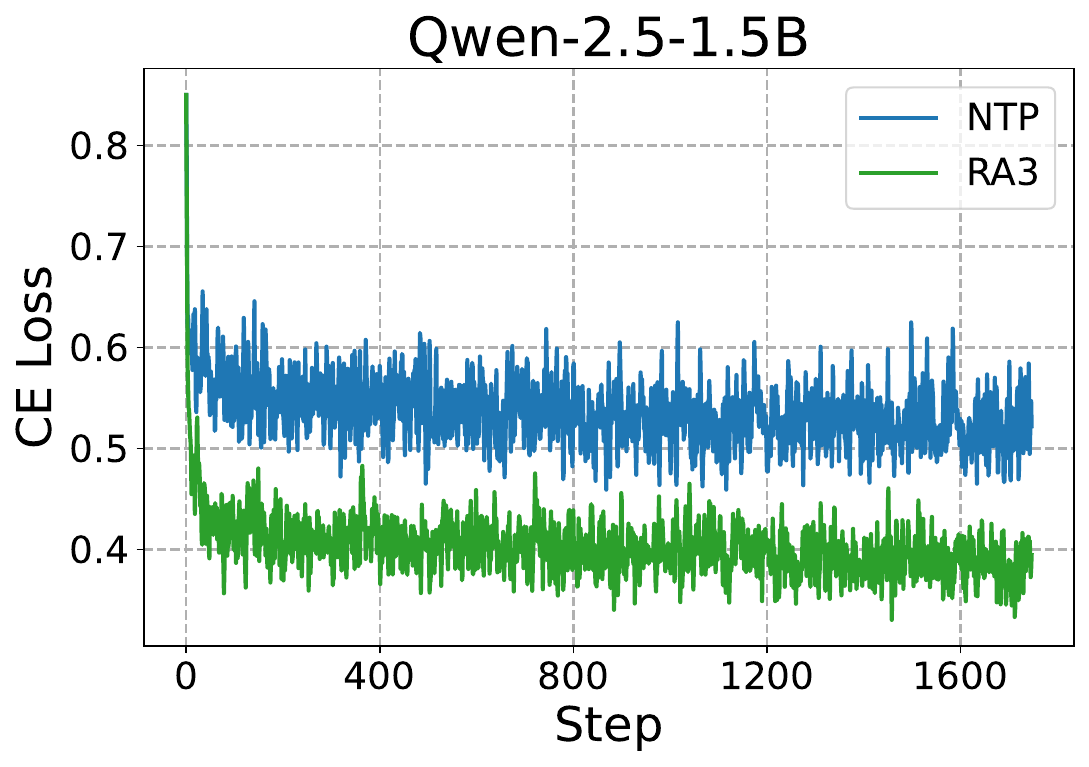}
    \end{minipage}
    \begin{minipage}[b]{0.325\linewidth}
        \centering
        \includegraphics[width=1\linewidth]{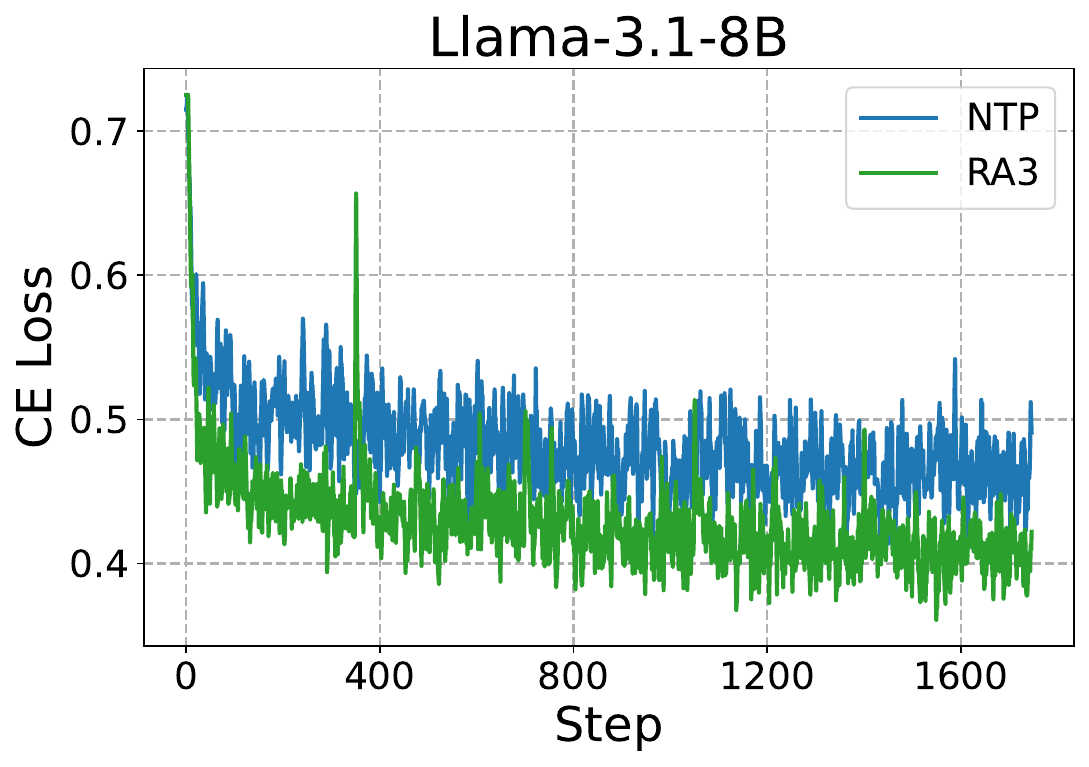}
    \end{minipage}
\vspace{-0.2cm}
\caption{Data bootstrapped with reasoning learned in the E step reduces the CE loss during the M step fine-tuning.}
\vspace{-0.2cm}
\label{fig_ce_loss}
\end{figure}

For the M step, an important metric is the cross-entropy (CE) loss, i.e., the negative log-likelihood of the next token. It can be observed from Figure \ref{fig_ce_loss} that fine-tuning on reasoning-bootstrapped data significantly accelerates learning. This supports our hypothesis on the mid-training data: while expert rollouts provide only primitive actions (raw code lines), hidden reasoning trajectories guide those decisions. Extracting such reasoning via RL makes learning easier and more efficient.

\begin{figure}[htbp]
    \centering
    \begin{minipage}[b]{0.32\linewidth}
        \centering
        \includegraphics[width=1.0\linewidth]{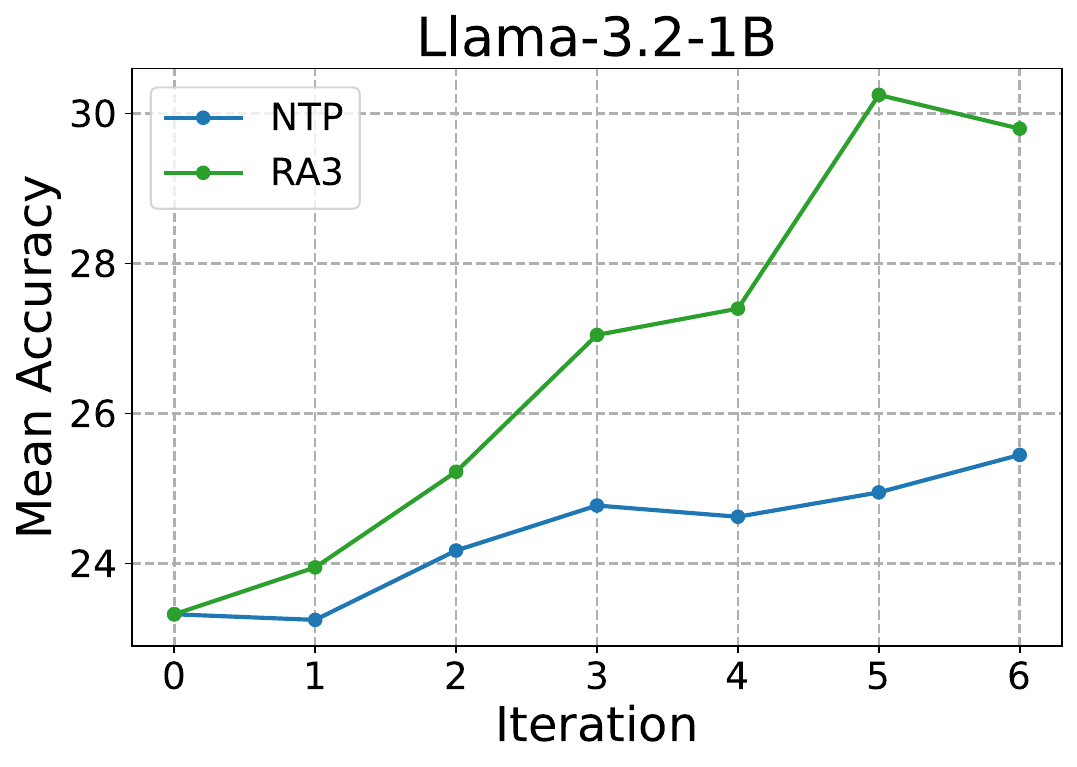}
    \end{minipage}
    \begin{minipage}[b]{0.32\linewidth}
        \centering
        \includegraphics[width=1.0\linewidth]{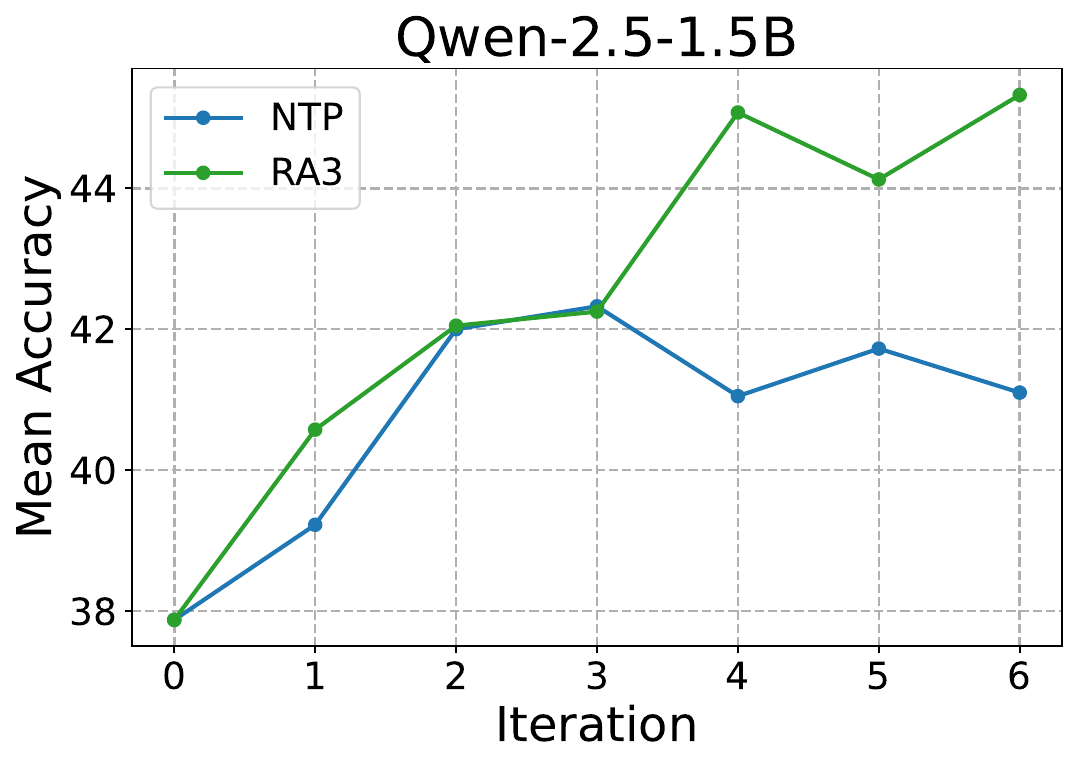}
    \end{minipage}
    \begin{minipage}[b]{0.32\linewidth}
        \centering
        \includegraphics[width=1\linewidth]{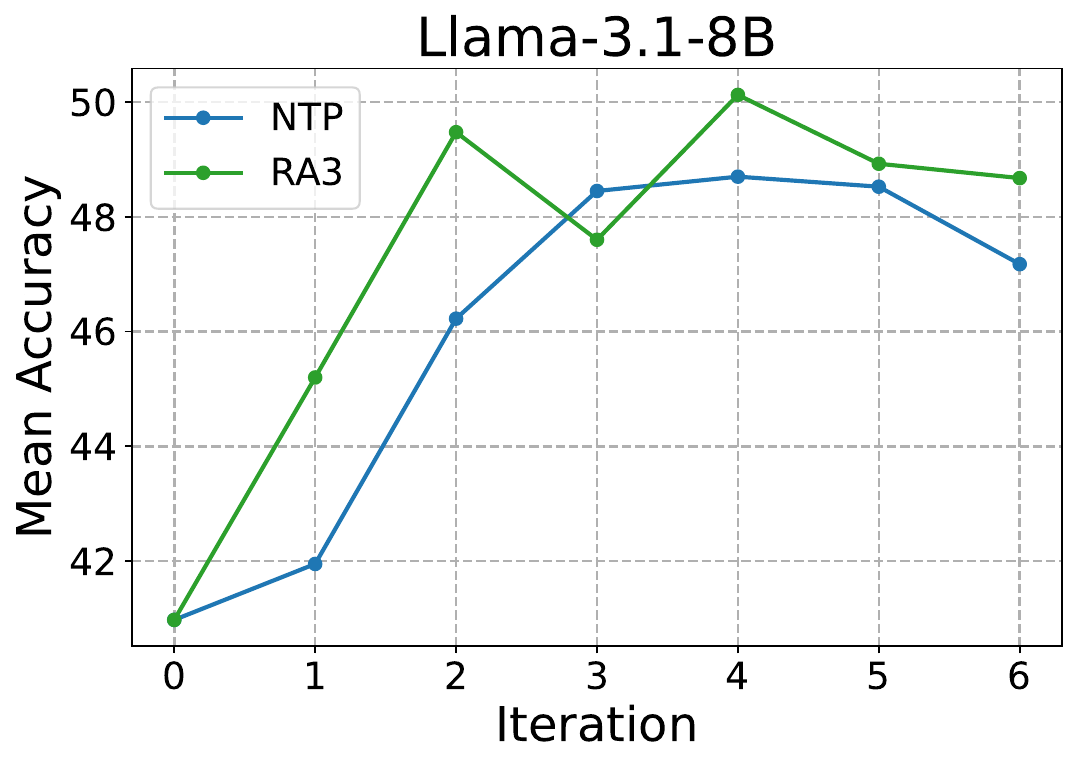}
    \end{minipage}
\vspace{-0.2cm}
\caption{Evaluation results during mid-training, with accuracies averaged across four benchmarks.}
\vspace{-0.1cm}
\label{fig_em_improve}
\end{figure}

We further evaluate the resulting models on the HumanEval \citep{chen2021evaluating}, MBPP \citep{austin2021program}, HumanEval+, and MBPP+ \citep{liu2023your} benchmarks using the BigCode evaluation harness \citep{bigcode-evaluation-harness}. We test the improvement of each EM iteration of RA3 by the average scores on these benchmarks, and compare with the NTP checkpoints that are trained on the same data. Figure \ref{fig_em_improve} tracks performance across EM iterations, showing that reasoning as action abstractions reduces CE loss and improves evaluation accuracy with fewer data compared to NTP.

Table \ref{tab_1} reports pass@k for the best NTP and RA3 checkpoints. RA3 consistently outperforms NTP, underscoring the generalization benefits of temporal action abstractions.

\vspace{-0.2cm}
\begin{table}[b]\small
\centering
\caption{Mid-training performance comparison: pass@k on HumanEval, MBPP, HumanEval+, and MBPP+.}
\label{tab:he-mbpp}
\setlength{\tabcolsep}{6.0pt}
\begin{tabular}{llcccccccccc}
\toprule
\multirow{2}{*}{Model} & \multirow{2}{*}{Method} & \multicolumn{2}{c}{HumanEval} & \multicolumn{2}{c}{MBPP} & \multicolumn{2}{c}{HumanEval+} & \multicolumn{2}{c}{MBPP+} & \multicolumn{2}{c}{Avg.} \\
\cmidrule(lr){3-4}\cmidrule(lr){5-6}\cmidrule(lr){7-8}\cmidrule(lr){9-10}\cmidrule(lr){11-12}
& & p@1 & p@5 & p@1 & p@5 & p@1 & p@5 & p@1 & p@5 & p@1 & p@5 \\
\midrule
\multirow{3}{*}{Llama 3.2 1B}
  & Base & 18.9 & 30.4 & 25.8 & 37.0 & 17.1 & 25.7 & 31.5 & 47.9 & 23.3 & 35.3 \\
  & NTP  & 21.3 & 34.1 & 27.8 & 45.8 & 17.7 & 29.5 & 34.4 & 51.8 & 25.3 & 40.3 \\
  & RA3 & \textbf{25.0} & \textbf{38.1} & \textbf{32.8} & \textbf{46.1} & \textbf{22.0} & \textbf{33.5} & \textbf{39.4} & \textbf{54.6}  & \textbf{29.8} & \textbf{43.1} \\
\midrule
\multirow{3}{*}{Qwen 2.5 1.5B}
  & Base & 37.2 & 53.7 & 38.6 & 56.5 & 32.3 & 48.1 & 43.4 & 61.0 & 37.9 & 54.8 \\
  & NTP  & 41.5 & 57.3 & 43.4 & 59.2 & 35.4 & 53.3 & 46.6 & 64.8 & 41.7 & 58.7 \\
  & RA3 & \textbf{48.2} & \textbf{63.6} & \textbf{45.8} & \textbf{61.3} & \textbf{42.7} & \textbf{59.0} &\textbf{49.7} & \textbf{64.9} & \textbf{46.6} & \textbf{62.2} \\
\midrule
\multirow{3}{*}{Llama 3.1 8B}
  & Base & 36.6 & 57.5 & 45.2 & 59.1 & 30.5 & 54.7 & 51.6 & 65.7 & 41.0 & 59.3 \\
  & NTP  & 48.2  & 62.0 & \textbf{48.6} & 62.9 & 42.7 & 60.1 & 51.1 & 67.4 & 47.7 & 63.1 \\
  & RA3 & \textbf{50.0} & \textbf{66.2} & 48.0 & \textbf{63.0} & \textbf{44.5} & \textbf{61.3} &\textbf{53.2} & \textbf{67.8} & \textbf{48.9}& \textbf{64.6}\\
\bottomrule
\end{tabular}
\label{tab_1}
\end{table}

\begin{figure}[h]
    \centering
    \includegraphics[width=1.0\linewidth]{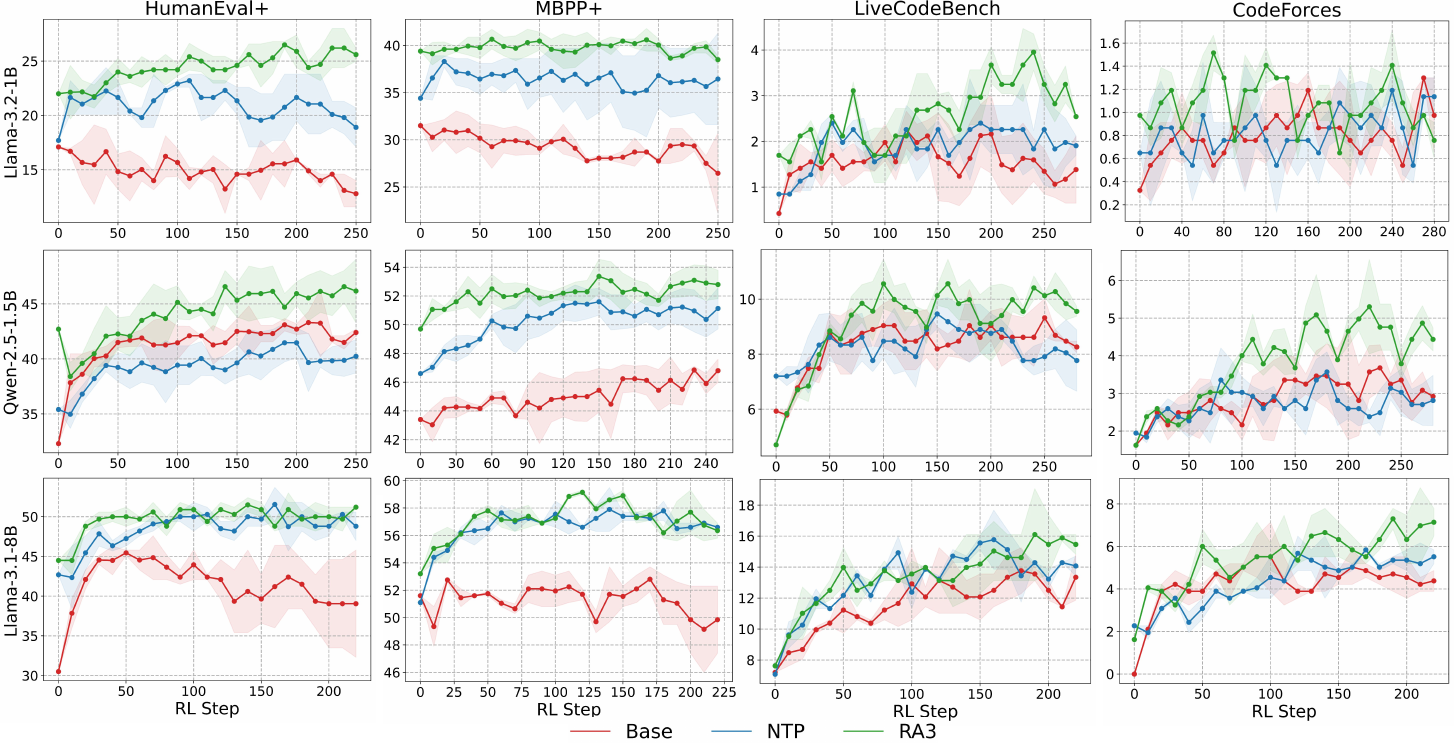}
    \vspace{-0.6cm}
\caption{RLVR evaluation results (mean and standard error across independent runs) of different mid-training algorithms.}
\vspace{-0.35cm}
\label{fig_rlvr}
\end{figure}

\paragraph{Post-Training RLVR Results.} We next study the impact of mid-training on post-training RL. We use GRPO as the default RLVR algorithm, and leverage the off-the-shelf DeepCoder codebase \citep{deepcoder2025}. We use the AReaL-boba-2-RL-Code dataset \citep{fu2025areal} for training, which contains $7.7$K data, filtered from TACO \citep{li2023taco}, \cite{2025synthetic1}, and LiveCodeBench \citep{jain2024livecodebench}. Instead of using the chat format and parsing the code within a block, we provide the function signature at the end of the instruction. This way, base models that do not support chat templates can directly complete the function body following the given signature. We run independent RLVR training with different random seeds ($3$ for small models and $2$ for the 8B model), and the evaluation results are reported in Figure \ref{fig_rlvr}.

We observe that mid-training significantly improves the RLVR performance compared to the base models. Besides, RA3 is able to learn more effectively during RLVR, in terms of both convergence speed and asymptotic performance, supporting our theoretical results in Section \ref{sec_analysis}.

\begin{figure}[h]
    \centering
    \begin{minipage}[b]{0.235\linewidth}
        \centering
        \includegraphics[width=1.0\linewidth]{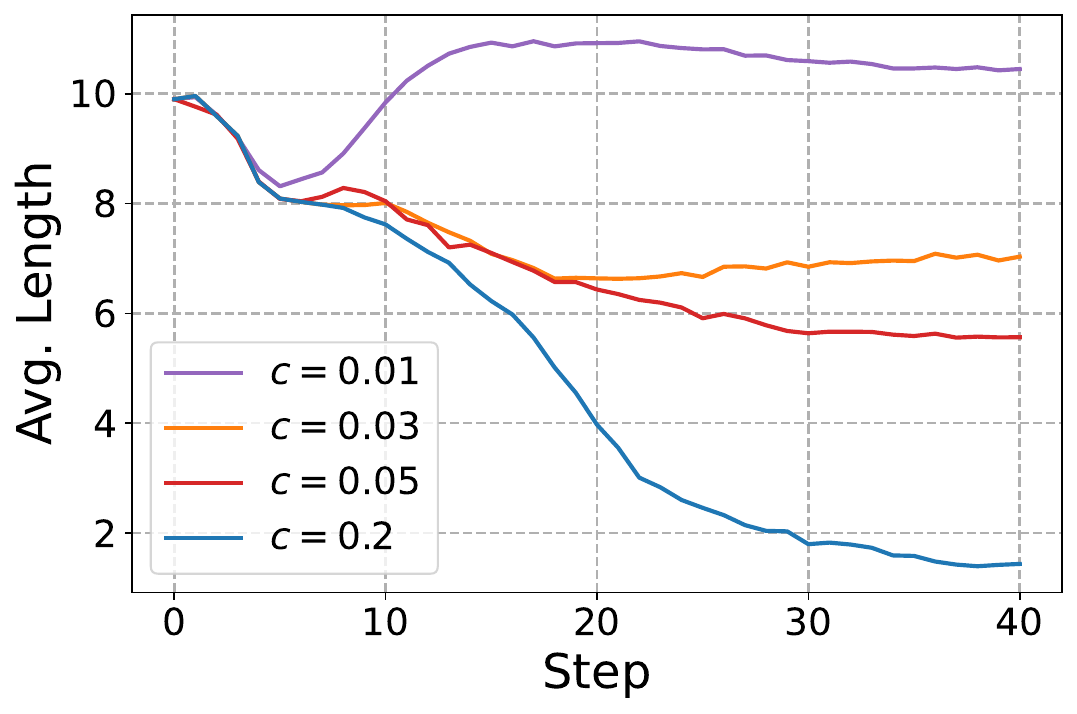}
    \end{minipage}
    \begin{minipage}[b]{0.245\linewidth}
        \centering
        \includegraphics[width=1.0\linewidth]{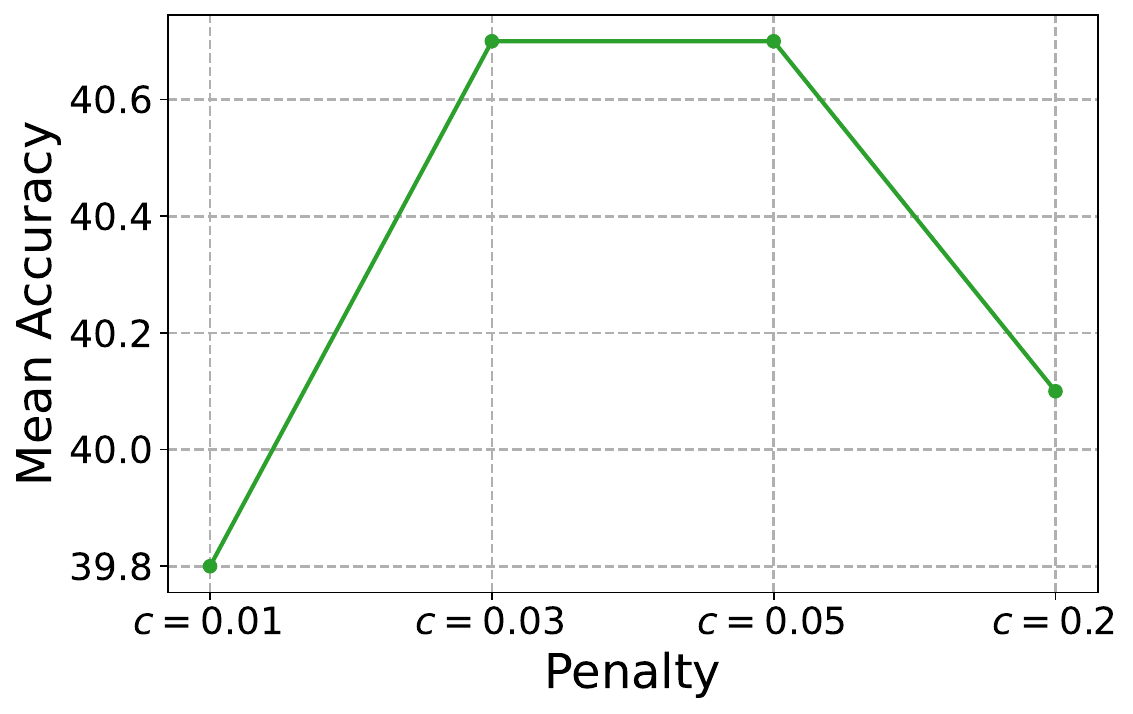}
    \end{minipage}
    \begin{minipage}[b]{0.24\linewidth}
        \centering
        \includegraphics[width=1.0\linewidth]{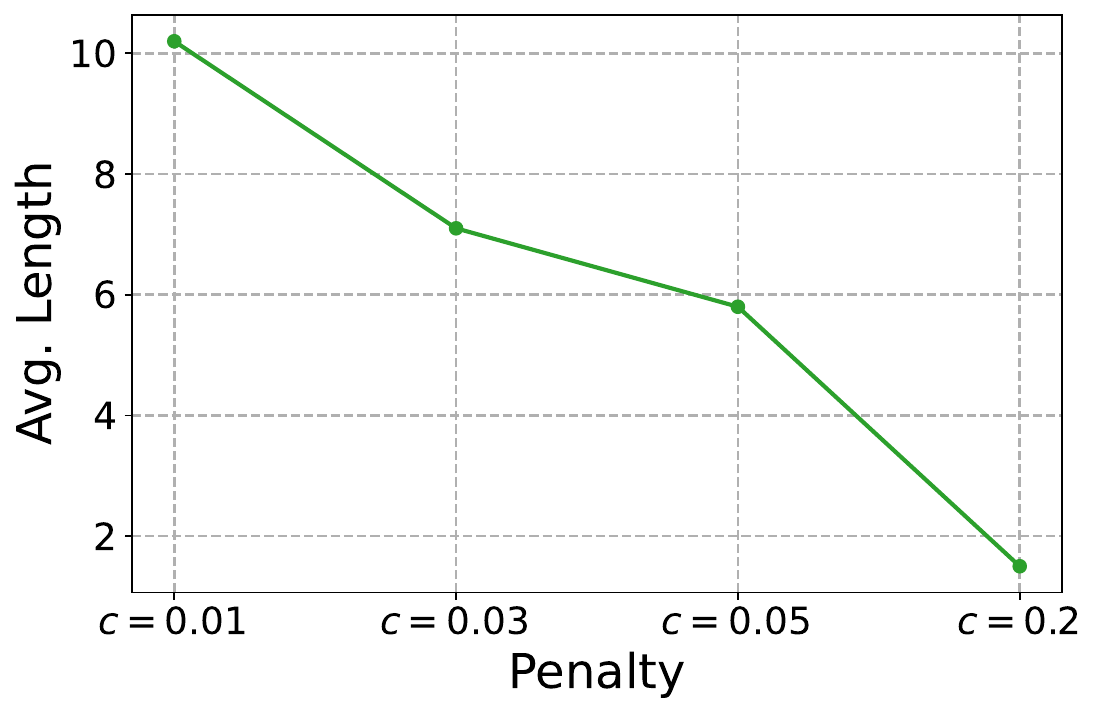}
    \end{minipage}
    \begin{minipage}[b]{0.26\linewidth}
        \centering
        \includegraphics[width=1\linewidth]{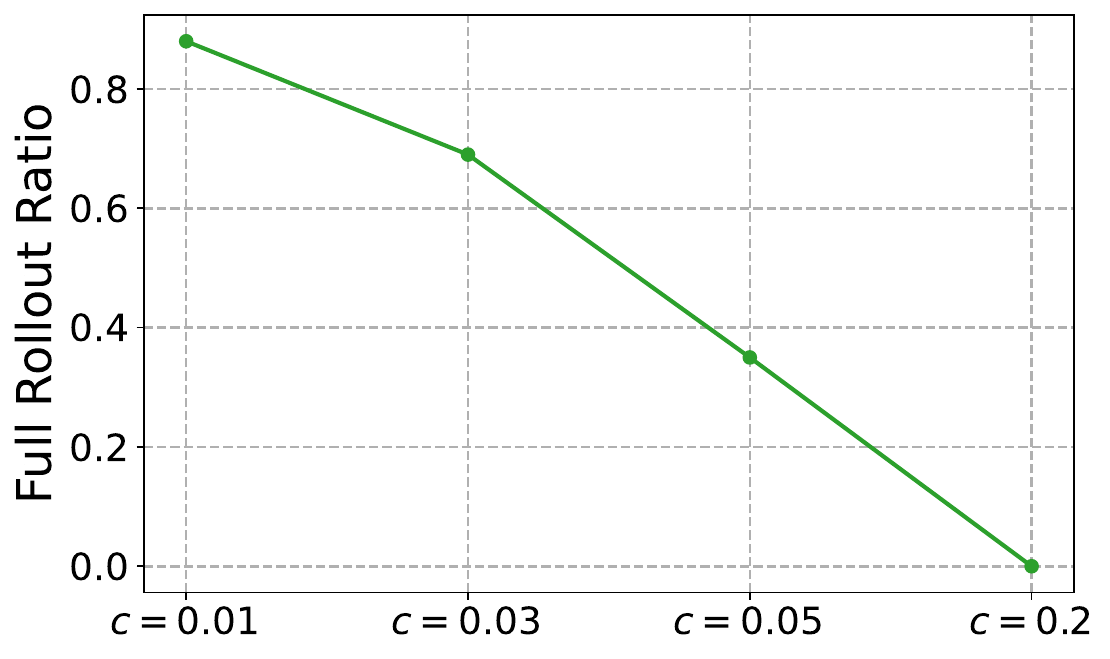}
    \end{minipage}
\vspace{-0.6cm}
\caption{Effect of penalty $c$ on (a) the behavior of the RL step, (b) the mean accuracy of HumanEval and MBPP, (c) the average length of $z$, and (d) the ratio of full rollout samples.}
\vspace{-0.3cm}
\label{fig_ablation}
\end{figure}

\paragraph{Ablation Study.} We investigate the role of the KL penalty $c$ when training the Qwen model. Recall that $c$ regulates the temporal consistency of the latent action $z$ by enforcing a threshold for when reasoning should occur. As shown in Figure \ref{fig_ablation}, a small $c$ causes the $q$-policy to generate redundant $z$ values at nearly every timestep. This behavior is expected: for most steps, there exists some $z_t \neq \text{\textless{}act\textgreater{}}$ that explains $a_{t+1}$ more effectively (i.e., yields higher log-likelihood RL reward) than the null action $z_t = \text{\textless{}act\textgreater{}}$. However, such behavior offers no advantage over primitive-action NTP, since neither the decision space nor the effective planning horizon is reduced. Moreover, the computational overhead of RA3 can be controlled through $c$: a large penalty $c=0.2$ requires that $\text{\textless{}think\textgreater{}}$ improves the expert log-likelihood by at least $0.2$ to be rewarded, an unlikely event. In this regime, the policy learns to output only $\text{\textless{}act\textgreater{}}$, and RA3 degenerates into NTP. Our default $c=0.05$ strikes a balance: reasoning is triggered at fewer than $40\%$ of lines, producing short rationales while maintaining efficiency.

\vspace{-0.15cm}

\section{Conclusion}
An effective mid-training stage should extract from expert demonstrations the subset of actions that are sufficient for all tasks and structure the decision space to enable fast RL convergence. In this paper, we present the first formal analysis of how mid-training design choices shape post-training RL. Our findings highlight two critical roles of mid-training: (1) efficiently pruning the action space, which forms a strong policy prior, and (2) accelerating subsequent RL convergence, which maximizes policy improvement through online interactions. Both insights favor learning in the space of temporally extended actions, rather than the large primitive action space of next-token prediction. Guided by this principle, we propose a novel mid-training algorithm based on the temporal variational bound. This method iteratively uncovers temporally-consistent reasoning trajectories via RL and fine-tunes on the resulting bootstrapped data. Experiments on code generation tasks confirm that reasoning as action abstraction enhances generalizability of the initial policy and improves post-training RLVR in both convergence speed and asymptotic performance.

\bibliographystyle{plainnat}
\bibliography{biblio}

\appendix
\section{Proofs}\label{app_proof}
\subsection{Proof of Lemma \ref{lemma_r}}
\begin{proof}
Since the policy $\pi$ selects actions from $\hat{\mathcal{Z}}$, it is an admissible policy in both $\mathcal{M}$ and $\mathcal{M}_{\hat{\mathcal{Z}}}$. Because the dynamics, rewards, and discount factors are identical for $\mathcal{M}$ and $\mathcal{M}_{\hat{\mathcal{Z}}}$, we have $V_{\mathcal{M}}^\pi =V_{\mathcal{M}_{\hat{Z}}}^\pi$. The results then follow from basic algebra.
\end{proof}

\subsection{Proof of Theorem \ref{thm_expert_size}}
Before the proof, we first define the subset of suboptimal actions.
\begin{definition}[Suboptimal Action Subset]\label{def_bad}
$\mathcal{Z}'$ is $(\epsilon, \sigma)$-suboptimal if $\EE_{\mathcal{M}\sim p(\mathcal{M})}[\mathmybb{1}(\Delta(\mathcal{M}, \mathcal{Z}')>\epsilon)]\geq\sigma$.
\end{definition}
\begin{proof}
For an $(\epsilon, \sigma)$-suboptimal action subset $\mathcal{Z}'$, according to Definition \ref{def_bad}, we know that 
\$
\Pr\limits_{\mathcal{M}\sim p(\mathcal{M})}[\Delta(\mathcal{M}, \mathcal{Z}')>\epsilon]\geq\sigma.
\$
Therefore, for a randomly drawn task $\mathcal{M}\sim p(\mathcal{M})$, the probability that $\mathcal{Z}'$ is not pruned away is no more than $1 - \sigma$.

With $|\mathcal{D}_E|$ independent expert demonstrations in mid-training, we reject $\mathcal{Z}'$ if $V_{\mathcal{M}^i}^*(s_0) \leq V_{\mathcal{M}^i_{\mathcal{Z}'}}^*(s_0) + \epsilon$ for any $i\in[1, |\mathcal{D}_E|]$. The probability that $\mathcal{Z}'$ is not pruned away during mid-training is thus no more than $(1 - \sigma)^{|\mathcal{D}_E|}$. We denote it as
\$
p\bigl(\text{survive}(\mathcal{Z}'\bigr)) \leq (1 - \sigma)^{|\mathcal{D}_E|}\leq e^{-\sigma|\mathcal{D}_E|},
\$
where the second inequality holds since $1+x\leq e^x$ for all real $x$.

Let $\mathcal{B} = \{\hat{\mathcal{Z}}\subseteq\mathcal{Z}: |\hat{\mathcal{Z}}|= |\overline{\mathcal{Z}}_\epsilon|\}$ be the set of all $|\overline{\mathcal{Z}}_\epsilon|$-size subsets.
We care about the event that there exist some suboptimal subsets of actions that are not pruned away, i.e., survive:
\$
p\Biggl(\bigcup_{\mathcal{Z}'\in\mathcal{B}} \text{survive}(\mathcal{Z}')\Biggr)\leq \sum_{\mathcal{Z}'\in\mathcal{B}} p\bigl(\text{survive}(\mathcal{Z}')\bigr)\leq \binom{|\mathcal{Z}|}{|\overline{\mathcal{Z}}_\epsilon|}e^{-\sigma|\mathcal{D}_E|} = \Theta\Bigl(|\mathcal{Z}|^{|\overline{\mathcal{Z}}_\epsilon|}\Bigr)e^{-\sigma|\mathcal{D}_E|},
\$
where the inequalities hold by applying a union bound over all candidate subsets.

Plugging in $|\mathcal{D}_E|=\Theta(|\overline{\mathcal{Z}}_\epsilon|\log(|\mathcal{Z}| / \delta) / \sigma)$ gives us
\$
p\Biggl(\bigcup_{\mathcal{Z}'\in\mathcal{B}} \text{survive}(\mathcal{Z}')\Biggr)\leq\delta,
\$
i.e., all the $(\epsilon, \sigma)$-suboptimal action subsets $\mathcal{Z}'$ are pruned away with probability at least $1-\delta$.

In other words, any action subset $\hat{\mathcal{Z}}$ after pruning is \textit{not} $(\epsilon, \sigma)$-suboptimal with high probability, i.e., it satisfies $\EE_{\mathcal{M}\sim p(\mathcal{M})}[\mathmybb{1}(\Delta(\mathcal{M}, \hat{\mathcal{Z}})>\epsilon)]<\sigma$, or equivalently $\EE_{\mathcal{M}\sim p(\mathcal{M})}[\mathmybb{1}(\Delta(\mathcal{M}, \hat{\mathcal{Z}})\leq\epsilon)]\geq1-\sigma$, with probability at least $1-\delta$.

Since the rewards are in $[0, 1]$, we know that $V_\mathcal{M}^{*}(s_0)\leq\sum_{t=0}^\infty \gamma^t = 1/(1-\gamma)$. Thus, $\Delta(\mathcal{M}, \hat{\mathcal{Z}})\leq 1/(1-\gamma)$.

Therefore, with probability at least $1-\delta$, the pruning error $\EE_{\mathcal{M}\sim p(\mathcal{M})}[\Delta(\mathcal{M}, \hat{\mathcal{Z}})] \leq \epsilon(1-\sigma)+\sigma/(1-\gamma)$.
\end{proof}

\subsection{Proof of Theorem \ref{thm_conv}}
\begin{proof}
We first define the operator $\mathcal{T}$ and $\mathcal{T}^\pi$ in the $\mathcal{Z}$ action space as
\$
(\mathcal{T}V)(s) = \max_{z\in\mathcal{Z}}\Bigl\{\EE_{\tau,s'}\bigl[R_\tau + \gamma^\tau V(s')\mid s, z\bigr]\Bigr\}, \quad (\mathcal{T}^\pi V)(s) = \EE_{z\sim\pi(\cdot\mid s)}\EE_{\tau,s'}\bigl[R_\tau + \gamma^\tau V(s')\mid s, z\bigr],
\$
where $p(s'|s, z) = \sum_{j}\gamma^j p(s', \tau=j | s, z)$, $p(s', \tau=j | s, z)$ is the joint probability of transitioning to $s'$ in $\tau$ steps after taking action $z$ at $s$, and $R_\tau = \sum_{k=1}^{\tau}\gamma^k R_{k}$.

For any functions $V_1$, $V_2$ and any state $s$, we have
\$
(\mathcal{T}V_1)(s) - (\mathcal{T}V_2)(s) &= \max_{z\in\mathcal{Z}}\Bigl\{\EE_{\tau,s'}\bigl[R_\tau + \gamma^\tau V_1(s')\mid s, z\bigr]\Bigr\} - \max_{z\in\mathcal{Z}}\Bigl\{\EE_{\tau,s'}\bigl[R_\tau + \gamma^\tau V_2(s')\mid s, z\bigr]\Bigr\}\\
&\leq \max_{z\in\mathcal{Z}}\EE\bigl[\gamma^\tau V_1(s') - V_2(s')\mid s, z\bigr],
\$
where the inequality holds since $\max f - \max g \leq \max(f - g)$ for any $f, g$.

Let $\left\|\cdot \right\|_\infty$ be the sup norm on functions $V: \mathcal{S}\mapsto\mathbb{R}$. Then
\$
(\mathcal{T}V_1)(s) - (\mathcal{T}V_2)(s) \leq\max_{z\in\mathcal{Z}}\Bigl\{\EE\bigl[\gamma^\tau\bigr]\cdot \left\|V_1 - V_2 \right\|_\infty\Bigr\}\leq\overline{\gamma}\left\|V_1 - V_2 \right\|_\infty.
\$
Since the above inequality holds for all $s\in\mathcal{S}$, taking the supremum over $s$ gives us $\left\|\mathcal{T}V_1 - \mathcal{T}V_2 \right\|_\infty\leq\overline{\gamma}\left\|V_1 - V_2 \right\|_\infty$. 
Therefore, $\mathcal{T}$ is a $\overline{\gamma}$-contraction on $(\mathbb{R}^{\mathcal{S}}, \left\|\cdot \right\|_\infty)$.

In policy iteration, since $\pi_{n+1}$ is greedy w.r.t. $V^{\pi_n}$, we have that
\$
V^{\pi_n} = \mathcal{T}_{\pi_n}V^{\pi_n}\leq \mathcal{T}V^{\pi_n} = \mathcal{T}_{\pi_{n+1}}V^{\pi_n}\leq V^{\pi_{n+1}}.
\$
By induction, we have $\mathcal{T}^n V^{\pi_0}\leq V^{\pi_n}\leq V^*$. Taking component-wise absolute values and then the maximum over the states, we get that
\$
\left\|V^* - V^{\pi_n} \right\|_\infty \leq \left\|V^* - \mathcal{T}^n V^{\pi_0}~\right\|_\infty = \left\|\mathcal{T}^n V^* - \mathcal{T}^n V^{\pi_0}~\right\|_\infty \leq \overline{\gamma}^n\left\|V^* - V^{\pi_0} \right\|_\infty.
\$
 
Next, we show that after every $\mathcal{O}(1/(1-\overline{\gamma}))$ iterations, policy iteration eliminates at least one suboptimal state-action pair. For notation convenience, we define the operator $(M_\pi V)(s) = \EE_{z\sim\pi(\cdot|s)}\EE_{\tau, s'}[\gamma^\tau V(s') | s, z]$. Then we may write $\mathcal{T}_{\pi}V = r_\pi + M_{\pi}V$ and $V^\pi = (I - M_{\pi})^{-1}r_\pi$. Thus, we obtain that
\$
V^* - \mathcal{T}_{\pi_n}V^*= (I - M_{\pi_n})(V^* - V^{\pi_n}) = (V^* - V^{\pi_n}) - M_{\pi_n}(V^* - V^{\pi_n}) \leq V^* - V^{\pi_n}.
\$

Taking the supremum gives us that
\$
\left\|V^* - \mathcal{T}_{\pi_n}V^* \right\|_\infty &\leq \left\|V^* - V^{\pi_n}\right\|_\infty\leq \overline{\gamma}^n\left\|V^* - V^{\pi_0}\right\|_\infty \\
&= \overline{\gamma}^n\left\|(I - M_{\pi_0})^{-1}(V^* - \mathcal{T}_{\pi_0}V^*) \right\|_\infty \leq \frac{\overline{\gamma}^n}{1 - \overline{\gamma}}\left\|V^* - \mathcal{T}_{\pi_0}V^* \right\|_\infty.
\$

Let $\overline{s}\in\mathcal{S}$ be the state that satisfies $V^*(\overline{s}) - \mathcal{T}_{\pi_0}V^*(\overline{s}) = \left\|V^* - \mathcal{T}_{\pi_0}V^* \right\|_\infty$. Then we have
\$
V^*(\overline{s}) - \mathcal{T}_{\pi_n}V^*(\overline{s})\leq \left\|V^* - \mathcal{T}_{\pi_n}V^* \right\|_\infty\leq \frac{\overline{\gamma}^n}{1 - \overline{\gamma}}\Bigl(V^*(\overline{s}) - \mathcal{T}_{\pi_0}V^*(\overline{s})\Bigr).
\$
Therefore, for some $n^*=\mathcal{O}(1/(1-\overline{\gamma}))$, when $n>n^*$, $\frac{\overline{\gamma}^n}{1 - \overline{\gamma}} < 1$. This indicates that
\$
\mathcal{T}_{\pi_n}V^*(\overline{s}) > \mathcal{T}_{\pi_0}V^*(\overline{s}).
\$
Therefore, $\pi_n(\overline{s})\neq \pi_0(\overline{s})$, i.e., at least one suboptimal state-action pair is eliminated by the policy iteration algorithm. Since there are $|\mathcal{S}||\hat{\mathcal{Z}}| - |\mathcal{S}|$ suboptimal state-action pairs in total, the policy achieves optimality after at most $\mathcal{O}((|\mathcal{S}||\hat{\mathcal{Z}}| - |\mathcal{S}|) / (1 - \overline{\gamma}))$ iterations.
\end{proof}

\subsection{Proof of Theorem \ref{thm_elbo}}
\begin{proof}
For the maximum likelihood objective, after introducing the sequence $z_{0:T}$ of latents, we have
\#\label{eq_elbo1}
\log p(a_{0:T}\mid s_{0:T}) &= \log\sum_{z_{0:T}}p(a_{0:T}, z_{0:T}\mid s_{0:T})\notag\\
&= \log\sum_{z_{0:T}}p(a_{0:T}, z_{0:T}\mid s_{0:T})\frac{q(z_{0:T}\mid s_{0:T})}{q(z_{0:T}\mid s_{0:T})}\notag\\
&= \log\EE_{z_{0:T}\sim q(\cdot\mid s_{0:T})}\biggl[\frac{p(a_{0:T}, z_{0:T}\mid s_{0:T})}{q(z_{0:T}\mid s_{0:T})}\biggr]\notag\\
&\geq \EE_{z_{0:T}\sim q(\cdot\mid s_{0:T})}\biggl[\log\frac{p(a_{0:T}, z_{0:T}\mid s_{0:T})}{q(z_{0:T}\mid s_{0:T})}\biggr],
\#
where the last inequality follows from Jensen's inequality.

From the probabilistic graphical model, we obtain that
\#\label{eq_elbo2}
q(z_{0:T}\mid s_{0:T}) &= q(z_0\mid s_{0:T})\prod_{t=1}^T q(z_t\mid s_{0:T}, z_{0:t-1})\notag\\
&= q(z_0\mid s_0)\prod_{t=1}^T q(z_t\mid s_t, z_{0:t-1}).
\#

Besides, we have from the Bayes rule that
\#\label{eq_elbo3}
p(a_{0:T}, z_{0:T}\mid s_{0:T}) &= p(a_0, z_0\mid s_{0:T})\prod_{t=1}^T p(z_t, a_t\mid s_{0:T}, z_{0:t-1}, a_{0:t-1})\notag\\
&= p(a_0, z_0\mid s_{0:T})\prod_{t=1}^T p(z_t\mid s_{0:T}, z_{0:t-1}, a_{0:t-1}) p(a_t\mid s_{0:T}, z_{0:t}, a_{0:t-1})\notag\\
&= p(a_0, z_0\mid s_0)\prod_{t=1}^T p(z_t\mid s_{t}, z_{0:t-1}) p(a_t\mid s_{t}, z_{t})\notag\\
&= p(z_0\mid s_0)\prod_{t=1}^T p(z_t\mid s_{t}, z_{0:t-1}) \prod_{t=0}^T p(a_t\mid s_{t}, z_{t}).
\#

Plugging \eqref{eq_elbo2} and \eqref{eq_elbo3} into \eqref{eq_elbo1} gives us
\$
\log p(a_{0:T}\mid s_{0:T}) &\geq \EE_{z_{0:T}\sim q(\cdot\mid s_{0:T})}\biggl[\log\frac{p(z_0\mid s_0)\prod_{t=1}^T p(z_t\mid s_{t}, z_{0:t-1}) \prod_{t=0}^T p(a_t\mid s_{t}, z_{t})}{q(z_0\mid s_0)\prod_{t=1}^T q(z_t\mid s_t, z_{0:t-1})}\biggr]\notag\\
&= \EE_{z_t\sim q(\cdot\mid s_t, z_{0:t-1})}\Biggl[\sum_{t=0}^T\log p(a_t\mid s_t, z_t) - \log\frac{q(z_t\mid s_t, z_{0:t-1})}{p(a_t\mid s_{t}, z_{t})}\Biggr]\notag\\
&= \EE_{z_t\sim q(\cdot\mid s_t, z_{0:t-1})}\Biggl[\sum_{t=0}^T\log p(a_t\mid s_t, z_t) - \mathcal{D}_{\text{KL}}\Bigl(q(z_t\mid s_t, z_{0:t-1}) \,\,||\,\, p(z_t\mid s_t, z_{t-1})\Bigr)\Biggr],
\$
where $p(z_t| s_t, z_{t-1})$ is the prior distribution of the latent $z_t$. 
\end{proof}

\subsection{Proof of Proposition \ref{prop_1}}
\begin{proof}
According to the definition of KL divergence, we have
\#\label{eq_remain1}
\mathcal{D}_{\text{KL}}\Bigl(q(z_t\mid s_t, z_{0:t-1}) \,\,||\,\, p(z_t\mid s_t, z_{0:t-1})\Bigr) &= q_{\text{act}}\log\frac{q_{\text{act}}}{\alpha} + \sum_{z_t\neq \langle\text{act}\rangle}q(z_t\mid s_t, z_{0:t-1})\log\frac{q(z_t\mid s_t, z_{0:t-1})}{p(z_t\mid s_t, z_{0:t-1})}\notag\\
&= q_{\text{act}}\log\frac{q_{\text{act}}}{\alpha} +\sum_{z_t\neq \langle\text{act}\rangle} q(z_t\mid s_t, z_{0:t-1})\log\frac{q(z_t\mid s_t,z_{0:t-1})}{(1-\alpha)U(z_t)}.
\#
We define $\Bar{q}(z_t | s_t, z_{0:t-1}) = q(z_t | s_t, z_{0:t-1}) / (1 - q_{\text{act}})$ for $z_t\neq \langle\text{act}\rangle$. Then 
\#\label{eq_remain2}
&\sum_{z_t\neq \langle\text{act}\rangle} q(z_t\mid s_t, z_{0:t-1})\log\frac{q(z_t\mid s_t,z_{0:t-1})}{(1-\alpha)U(z_t)} \notag\\
&\qquad= (1 - q_{\text{act}})\Biggl[\sum_{z_t\neq \langle\text{act}\rangle} \Bar{q}(z_t\mid s_t, z_{0:t-1})\log\Bar{q}(z_t\mid s_t, z_{0:t-1})+\log\frac{1-q_{\text{act}}}{1-\alpha}-\EE_{\Bar{q}}\log U(z_t)\Biggr].
\#
From the KL divergence between Bernoulli distributions, we obtain that
\$
\mathcal{D}_{\text{KL}}\Bigl(\text{Bern}(q_{\text{act}}) \,\,||\,\, \text{Bern}(\alpha)\Bigl) = q_{\text{act}}\log\frac{q_{\text{act}}}{\alpha} + (1 - q_{\text{act}})\log\frac{1-q_{\text{act}}}{1-\alpha}.
\$

Besides, we have
\$
\sum_{z_t\neq \langle\text{act}\rangle} \Bar{q}(z_t\mid s_t, z_{0:t-1})\log\Bar{q}(z_t\mid s_t, z_{0:t-1})-\EE_{\Bar{q}}\log U(z_t) = \mathcal{D}_{\text{KL}}\Bigl(\Bar{q}(z_t \mid s_t, z_{0:t-1}) \,\,||\,\, U(z_t)\Bigl).
\$

By plugging the above two KL divergence terms into \eqref{eq_remain1} and \eqref{eq_remain2}, we obtain
\$
&\mathcal{D}_{\text{KL}}\Bigl(q(z_t\mid s_t, z_{0:t-1}) \,\,||\,\, p(z_t\mid s_t, z_{0:t-1})\Bigr) \\
&\quad=\mathcal{D}_{\text{KL}}\Bigl(\text{Bern}(q_{\text{act}}) \,\,||\,\, \text{Bern}(\alpha)\Bigl) - (1 - q_{\text{act}})\mathcal{D}_{\text{KL}}\Bigl(\Bar{q}(z_t \mid s_t, z_{0:t-1}) \,\,||\,\, U(z_t)\Bigl)\\
&\quad= \mathcal{D}_{\text{KL}}\Bigl(\text{Bern}(q_{\text{act}}) \,\,||\,\, \text{Bern}(\alpha)\Bigl) - \sum_{z_t\neq \langle\text{act}\rangle} q(z_t\mid s_t, z_{0:t-1})\log\frac{q(z_t\mid s_t, z_{0:t-1})}{U(z_t)(1 - q_{\text{act}})}\\
&\quad= \mathcal{D}_{\text{KL}}\Bigl(\text{Bern}(q_{\text{act}}) \,\,||\,\, \text{Bern}(\alpha)\Bigl) - \sum_{z_t\neq \langle\text{act}\rangle} q(z_t\mid s_t, z_{0:t-1})\log q(z_t\mid s_t, z_{0:t-1}) + (1 - q_{\text{act}})\log(1 - q_{\text{act}}) + \text{const}\\
&\quad=\mathcal{D}_{\text{KL}}\Bigl(\text{Bern}(q_{\text{act}}) \,\,||\,\, \text{Bern}(\alpha)\Bigl) - \mathcal{H}\Bigl(q(z_t\mid s_t, z_{0:t-1})\Bigr) + \text{const},
\$
where the constant comes from the uniform prior term that appears when expanding the log and can be omitted. This gives us the desired result.
\end{proof}

\end{document}